\documentclass[10pt,twocolumn,letterpaper]{article}

\usepackage[pagenumbers]{cvpr} 

\usepackage{graphicx}
\usepackage{amsmath}
\usepackage{amssymb}
\usepackage{booktabs}
\usepackage{siunitx}
\usepackage[accsupp]{axessibility}

\usepackage[pagebackref,breaklinks,colorlinks]{hyperref}

\usepackage[capitalize]{cleveref}
\crefname{section}{Sec.}{Secs.}
\Crefname{section}{Section}{Sections}
\Crefname{table}{Table}{Tables}
\crefname{table}{Tab.}{Tabs.}

\def\cvprPaperID{21}
\def\confName{CVPR}
\def\confYear{2023}

\makeatletter
\@namedef{ver@everyshi.sty}{}
\makeatother
\usepackage{tikz}
\usepackage{textcomp}
\usepackage{lipsum}
\newcommand\copyrighttext{
	\footnotesize \textcopyright 2023 IEEE. Personal use of this material is permitted. Permission from IEEE must be obtained for all other uses, in any current or future media, including reprinting/republishing this material for advertising or promotional purposes, creating new collective works, for resale or redistribution to servers or lists, or reuse of any copyrighted component of this work in other works.
}
\newcommand\copyrightnotice{%
	\begin{tikzpicture}[remember picture,overlay]
		\node[anchor=south,yshift=10pt, xshift=10pt] at (current page.south) {\fbox{\parbox{\dimexpr\textwidth-\fboxsep-\fboxrule\relax}{\copyrighttext}}};
	\end{tikzpicture}%
}

\begin{document}
	
	\title{RadarGNN: Transformation Invariant Graph Neural Network for Radar-based Perception}
	
	\author{Felix Fent,
	Philipp Bauerschmidt and
	Markus Lienkamp\\
	Technical University of Munich, Germany\\
	School of Engineering and Design\\
	Institute of Automotive Technology \\
	{\tt\small felix.fent@tum.de}
}
\maketitle
\copyrightnotice

\begin{abstract}
	
	A reliable perception has to be robust against challenging environmental conditions. Therefore, recent efforts focused on the use of radar sensors in addition to camera and lidar sensors for perception applications. However, the sparsity of radar point clouds and the poor data availability remain challenging for current perception methods. To address these challenges, a novel graph neural network is proposed that does not just use the information of the points themselves but also the relationships between the points. The model is designed to consider both point features and point-pair features, embedded in the edges of the graph. Furthermore, a general approach for achieving transformation invariance is proposed which is robust against unseen scenarios and also counteracts the limited data availability. The transformation invariance is achieved by an invariant data representation rather than an invariant model architecture, making it applicable to other methods. The proposed RadarGNN model outperforms all previous methods on the RadarScenes dataset. In addition, the effects of different invariances on the object detection and semantic segmentation quality are investigated. The code is made available as open-source software under \href{https://github.com/TUMFTM/RadarGNN}{https://github.com/TUMFTM/RadarGNN}.
	
\end{abstract}

\section{Introduction}
\label{sec:intro}

\begin{figure}[t]
	\centering
	\begin{subfigure}{\linewidth}
		\includegraphics[width=\linewidth]{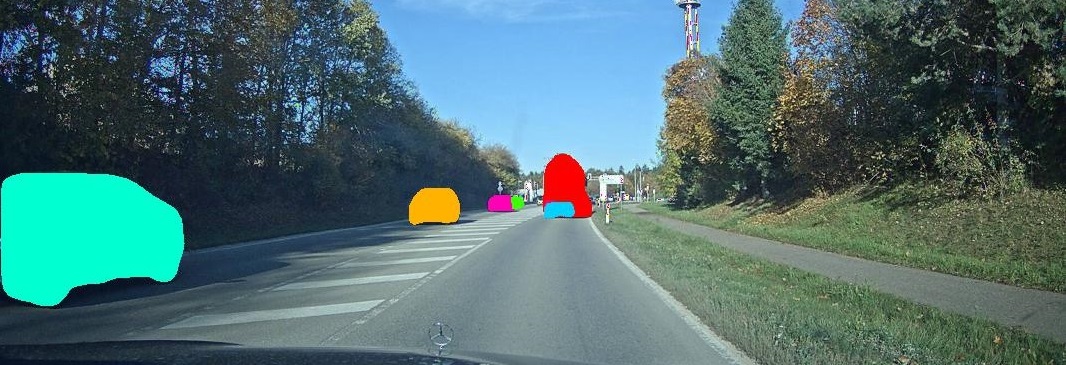}
		\caption{Example scene}
		\label{fig:intro-a}
	\end{subfigure}
	\hfill
	\begin{subfigure}{0.49\linewidth}
		\includegraphics[width=\linewidth]{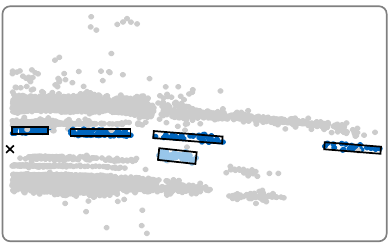}
		\caption{Ground truth}
		\label{fig:intro-b}
	\end{subfigure}
	\hfill
	\begin{subfigure}{0.49\linewidth}
		\includegraphics[width=\linewidth]{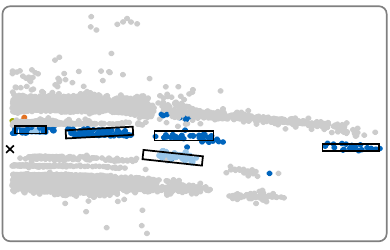}
		\caption{Model prediction}
		\label{fig:intro-c}
	\end{subfigure}
	\caption{Example scenario a) of the RadarScenes dataset~\cite{Schumann.2021} and its corresponding radar point cloud data in the bird's eye view. The annotated ground truth data is shown in b), while the model prediction for object classes and bounding boxes is given in c).}
	\label{fig:intro}
\end{figure}

Autonomous vehicles rely on an accurate representation and understanding of their environment. To achieve this, even under severe weather conditions, the perception has to be robust against changing environmental conditions. However, current perception systems rely mainly on data from camera or light detection and ranging (lidar) sensors, which are negatively affected by certain environmental conditions~\cite{Yoneda.2019}. The perception capability of both sensor types is for example reduced by fog or rain and camera sensors are dependent on an external light source limiting their usability in the dark~\cite{Yoneda.2019}. As a result of these limitations, research has focused on integrating radio detection and ranging (radar) sensors into perception systems.

Even if radar data is mostly unaffected by adverse environmental conditions~\cite{Yoneda.2019}, the detection quality of radar-based systems cannot yet compete with state-of-the-art image or lidar-based perception methods~\cite{Caesar.2020}. While there are several reasons for this discrepancy, the two major challenges of radar-based perception are the limited availability of annotated radar data, and the sparsity of radar point cloud data~\cite{Scheiner.2021, Svenningsson.2021, Zhou.2022}.

Leveraging the sparse information available, a graph neural network (GNN) is proposed that will not just utilize the information encoded in the points but also the relationships between the points. As shown in \cref{fig:intro}, an object is characterized by multiple radar points, which is why the relationship between the points is important to identify and differentiate between objects. In addition, GNNs can operate on unstructured and unordered input data, eliminating the need for data discretization (voxelization) and its associated loss of information~\cite{Shi.2020}. Therefore, all the information of sparse radar point clouds can be used without losing their structural information.

To counteract the limited data availability, a general approach for incorporating invariances into the perception pipeline is proposed. Building upon the success of translation invariant convolution operations, a method is proposed for creating a translation and rotation invariant perception pipeline, leading to better generalization and improved perception quality.

The proposed method was evaluated on the RadarScenes dataset~\cite{Schumann.2021} and outperforms all previous methods for bounding box prediction as well as semantic segmentation. In summary, the contributions of this paper are:

\begin{itemize}
	\item A novel GNN model for radar-based multi-class object detection and semantic segmentation.
	\item A general approach for transformation invariant object detection and semantic segmentation.
	\item A new state of the art for object detection and semantic segmentation on the RadarScenes dataset.
\end{itemize}

\section{Related Work}
\label{sec:sota}

State-of-the-art radar-based object detection methods rely on deep neural networks (DNN) to detect objects within the provided radar point clouds. Even if radar data can also be processed with more conventional methods~\cite{Scheiner.2019, Schumann.2017, Schumann.2018.1} and object detection can be performed at different data abstraction levels~\cite{Meyer.2021, Palffy.2020, Rebut.2022}, DNNs applied to radar point clouds achieve the best results.

\subsection{Radar Datasets}

The performance of data-driven perception methods is to a great extend dependent on the underlying dataset. Therefore, the selection of an appropriate dataset is essential for successful model training and meaningful evaluation. However, since most popular perception datasets, such as KITTI~\cite{Geiger.2012} or the Waymo Open Dataset~\cite{Sun.2020}, do not include annotated radar data, special emphasis is placed on radar-oriented datasets.

Of these, the Dense~\cite{Bijelic.2020}, PixSet~\cite{Deziel.2021} and Zendar~\cite{Mostajabi.2020} datasets provide annotated two-dimensional radar data, but the spatial resolution of the deployed radar sensors as well as the extend of the dataset is comparatively small. In contrast, the Oxford Radar RobotCar~\cite{Barnes.2020}, MulRan~\cite{Kim.2020} and RADIATE~\cite{Sheeny.2021} datasets utilize high resolution spinning radar sensors which are not representative of currently deployed automotive radar sensors. The nuScenes~\cite{Caesar.2020} dataset includes radar data, but multiple authors~\cite{Engels.2021, Nobis.2021, Schumann.2021, Zhou.2022} have criticized the radar data quality of the nuScenes dataset because of its sparsity, limited feature resolution and errors within the radar domain. In consequence, the RadarScenes~\cite{Schumann.2021} dataset is chosen for this work.

The RadarScenes~\cite{Schumann.2021} dataset includes point-wise annotated radar data of moving objects assigned to eleven different categories. The dataset comprises the data of four series production automotive radar sensors and contains more than four hours of driving data. The radar points are represented by their spatial coordinates ($x, y$), target velocities ($v_x, v_y$), radar cross section ($rcs$) and a timestamp ($t$). Currently, most comparative results for radar-based perception are reported based on the RadarScenes dataset, even if it does not provide ground truth bounding boxes and only considers moving objects for ground truth annotations.

\subsection{Point Cloud Object Detection}

The approaches used to detect objects within point clouds, using deep neural networks, can be divided into three major groups: point-based, grid-based and graph-based methods. In addition to these three general concepts, hybrid methods can be designed by combining different approaches.

Point-based approaches operate on the input point clouds directly, without the need for any preceding data transformations. Therefore, all information and the structural integrity of the point cloud is preserved. Utilizing this method, Schumann \etal~\cite{Schumann.2018.2} performed a semantic segmentation on a proprietary radar dataset and later extended their approach to develop an instance segmentation model for radar data~\cite{Scheiner.2021}. Building upon this, Nobis \etal~\cite{Nobis.2021} developed a point-based recurrent neural network (RNN) to realize semantic segmentation on nuScenes radar data. Nevertheless, point-based approaches cannot consider individual relationships between points, even if the structure of local groups can be taken into account~\cite{Shi.2020}.

Grid-based methods map the point clouds to a structured grid representation by a discretization (voxelization) of the underlying space. Based on this data structure, conventional convolutional neural networks (CNN) can be applied to accomplish different computer vision tasks. Using this approach, Schumann \etal~\cite{Schumann.2020} developed an autoencoder network to perform a semantic segmentation on radar point clouds, originating from a proprietary dataset. Scheiner \etal~\cite{Scheiner.2021} applied a YOLOv3~\cite{Redmon.2018} detector to a bird's-eye view (BEV) grid representation of radar point clouds and achieved state-of-the-art object detection results on the RadarScenes~\cite{Schumann.2021} dataset. However, the preceding data transformation leads to a loss of information and a sparse data representation.

Graph-based methods construct a graph from the input point cloud to operate on and can be categorized into convolutional~\cite{Kipf.2016}, attentional~\cite{Velickovic.2018} and message passing~\cite{Gilmer.2017} neural network types~\cite{Bronstein.2021}. For graph neural networks, the points are used as nodes within the graph, preserving the structural information of the point cloud, and the relationships between the points are modeled as edges in the graph~\cite{Bronstein.2017}. This methods was first used by Shi and Rajkumar~\cite{Shi.2020} to implement a GNN for object detection on lidar point clouds. So far, GNNs have only been used once by Svenningsson \etal~\cite{Svenningsson.2021} to realize a graph-based object detection on radar point clouds. However, their method was limited to a graph convolutional layer formulation, the effects of invariances where not investigated and their approach was limited to the detection of cars within the nuScenes dataset. Therefore, to the best of our knowledge, GNNs have never previously been used to accomplish a multi-class object detection task on radar point cloud data.

Additionally, hybrid methods can be used to combine several of the above mentioned techniques. Scheiner \etal~\cite{Scheiner.2021} compared multiple different hybrid methods on the RadarScenes~\cite{Schumann.2021} dataset and most recently, Palffy \etal~\cite{Palffy.2022} evaluated a hybrid model architecture on the newly published View-of-Delft~\cite{Palffy.2022} dataset. However, most hybrid methods~\cite{Tilly.2020, Scheiner.2021, Palffy.2022} rely on a grid-based approach for the final object detection and are therefore subject to similar disadvantages, as mentioned above.

\subsection{Transformation Invariance}

The great successes of convolutional neural networks over their predecessors was mainly justified by the translation invariant property of the convolution operation~\cite[pp.~335--339]{Goodfellow.2016}. A function is considered translation invariant if its output remains unchanged after a translation of its inputs~\cite{Bronstein.2017}.

Achieving an invariant object detection model is of great interest because the model should be robust to unseen scenarios, where objects can occur in different locations. Moreover, the model should also be applicable to sensors mounted in different positions. To accomplish this property, the following three methods have been used in the literature: data augmentation, invariant data representation and invariant model architectures.

Data augmentation is used to extend the training dataset by modified (e.g. translated or rotated) copies of the original data and relies on the model learning the aspired invariances during the training process~\cite{Benton.2020}. This technique is commonly used to support the generalization of the model and is applied in many of the above mentioned methods~\cite{Scheiner.2021, Schumann.2018.2, Svenningsson.2021}. However, this method does not ensure an invariant model after the training process.

Invariant data representation describes the restriction of the input features to those that are invariant to certain transformations~\cite[p. 23 ff.]{Mundy.1992}. In the Euclidean space, for example, distance is an invariant quantity, which is unaffected by all rigid transformations~\cite[p. 12]{Mundy.1992}. Only representing the data by such quantities results in an overall invariant data representation. However, limiting the input data to invariant features results in a loss of global context (e.g. absolute vs. relative coordinates).

Invariant model architectures are designed in such a way that the output of the model remains unchanged regardless of a transformation applied to the model input. Such an invariant model architecture was used by Svenningsson \etal~\cite{Svenningsson.2021} to design a translation invariant GNN for radar-based object detection. However, incorporating invariances in the model architecture restricts the architectural design to certain (mathematically symmetric) operations~\cite{Qi.2017} and makes the investigation of the effects of different invariances complicated.

\section{RadarGNN}
\label{sec:radargnn}

\begin{figure*}[t]
	\centering
	\includegraphics[width=\textwidth]{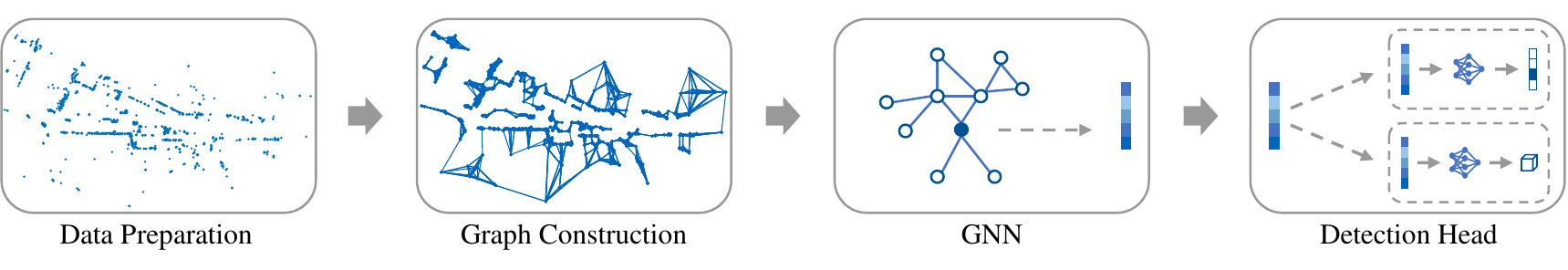}
	\caption{Model overview from point cloud processing on the left, through graph construction and GNN feature extraction, up to the object detection and semantic segmentation on the right.}
	\label{fig:overview}
\end{figure*}

This section describes the proposed RadarGNN model for radar-based object detection and semantic segmentation. The object detection pipeline is shown in \cref{fig:overview} and consists of four major steps: data preparation, graph construction, the graph neural network and the detection heads. The main idea behind this method is the achievement of transformation invariances not by using the model architecture itself but by creating an invariant data representation. To achieve this, three things are required: transformation invariant bounding boxes, a graph construction method for invariant input features and a generalization of the GNN layer to consider edge features.

\subsection{Data Preparation}

The purpose of the data preparation is to create a similar database to that used in previous research and to enable the training of a translation and rotation invariant perception model. This step is important for preserving the comparability of the results and affects not just the model inputs but also the target value determination.

The preparation of the RadarScenes data is based on the implementation of Scheiner \etal~\cite{Scheiner.2021}, who serves as a benchmark for this study. Using this approach, the model input is given by an accumulation of radar point clouds within a time period of \SI{500}{\milli\second}. The resulting point cloud is then cropped to an area of \SI{100}{\meter} times \SI{100}{\meter} in front of the vehicle's rear axis, as shown in~\cite[Fig. 2]{Scheiner.2021}. Since the RadarScenes dataset does not include ground truth bounding boxes, they are created as minimum enclosing rectangles in bird's-eye view, including all points belonging to the same instance. Furthermore, the original eleven instance categories are mapped to five major object classes and finally, the overall dataset is split into training (\SI{64}{\percent}), validation (\SI{16}{\percent}) and test (\SI{20}{\percent}) sets.

The investigation of the effects of different invariances on the detection quality of the model requires different bounding box definitions. The existing absolute bounding box definition can be used to train a non-invariant baseline model. An absolute bounding box is defined as a tuple $(x, y, w, l, \theta)$ consisting of the box center coordinates~$x, y$, the box dimensions~$w, l$ and the yaw angle~$\theta$.

The training of a translation invariant model requires a translation invariant bounding box definition. To achieve this, the bounding box is no longer defined by its absolute center coordinates but by the relative position to its associated radar point $p_0$. Therefore, a translation invariant bounding box is given by a tuple $(dx, dy, w, l, \theta)$, with a relative translation~$dx, dy$ between the box center and the radar point $p_0$ it belongs to.

\begin{figure}[t]
	\centering
	\includegraphics[width=\linewidth]{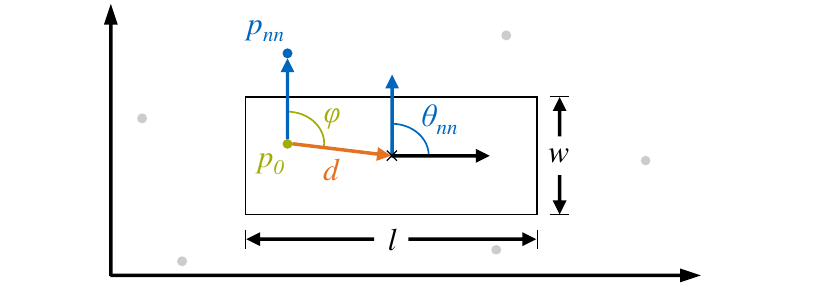}
	\caption{Definition of the translation and rotation invariant bounding box in regard to the radar point $p_0$ and the reference point $p_{nn}$. The box is defined by its extend ($w, l$), position ($d, \varphi$) and orientation ($\theta_{nn}$) in the bird's-eye view.}
	\label{fig:bounding_box}
\end{figure}

The bounding box definition for the training of a translation and rotation invariant model requires the addition of a second reference point $p_{nn}$. On this basis, a translation and rotation invariant bounding box is defined as tuple $(d, \varphi, w, l, \theta_{nn})$. Here, the distance $d$ is the distance between the reference point $p_0$ and the bounding box center. The angle $\varphi$ represents the angle between the vector from the reference point $p_0$ to its nearest neighbor $p_{nn}$ and the vector to the bounding box center. Finally, the angle $\theta_{nn}$ corresponds to the angle between the directional vector of the bounding box and the vector from the reference point $p_0$ to its nearest neighbor $p_{nn}$. A graphical representation of the translation and rotation invariant bounding box definition is given in \cref{fig:bounding_box}.

\subsection{Graph Construction}

The graph construction module transforms the initial radar point cloud into a graph representation and ensures a transformation invariant input data representation. The proposed method comprises of three major steps: the node feature transformation, the edge generation and a matrix transformation.

The node feature transformation maps the original point cloud~$\mathcal{P}$, formally defined as finite set~$\mathcal{P}$ of $n \in \mathbb{N}$ vectors~$p_{i} \in \mathbb{R}^{d}$ with $i=1, ..., n$, to a set of nodes~$\mathcal{V} = \{\nu_{0}, ..., \nu_{n} | \nu \in \mathbb{R}^{d_{\nu}}\}$. During this transformation process, the number of points is preserved while the features are transformed. This is important to ensure the invariance of the data representation, while preserving the structure of the original point cloud. The key functionality, however, is the selection of the node features, which determines the invariances of the data representation.

In the non-invariant baseline configuration, the nodes are defined by their absolute spatial coordinates ($x$, $y$), their velocity vectors ($v_x$, $v_y$), the values of the radar cross section~$rcs$ and the associated timestamps~$t$. To achieve a translation invariant data representation, the absolute spatial coordinates are omitted and the encoding of the structural information is subject to the edge generation. Accomplishing a translation and rotation invariant representation further requires the reduction of the velocity information to the Euclidean norm of the velocity vector~$v$. In addition, all nodes hold the information about the connectivity degree~$\text{c}$ (the number of associated edges). This process shows that the achievement of certain invariances results in a loss of global context, where the edge generation is an attempt at compensating for this.

The edge generation encodes relationships between points in the form of edges and edge features. The set of edges can be formally defined as $\mathcal{E} \subseteq \{(u, \nu) | (u, \nu)\in\mathcal{V}^{2}, u \neq \nu\}$, where every edge $\varepsilon \in \mathcal{E}$ can be associated with an edge feature vector $e: \mathcal{E} \mapsto \mathbb{R}^{d_{\varepsilon}}$. The final Graph is then defined by the tuple of nodes and edges $\mathcal{G} = (\mathcal{V}, \mathcal{E})$.

The edges of the graph are created by a k-nearest neighbors algorithm, where the number of neighbors k is determined empathically and represents a trade-off between performance and computational resources. As a result, every node of the graph is connected to its twenty nearest neighbors.

The edge features are generated in consideration of the desired invariances and with the aim of encoding relationship information between points. However, for the non-invariant baseline configuration, no edge features are generated but all information is contained within the node features. The translation invariant data representation requires the neglection of the absolute coordinates, instead the position relative to the neighboring points $dx$ and $dy$ is encoded in the edge features to preserve the spatial information. To create a rotation and translation invariant data representation, an enhanced set of point-pair features, inspired by Drost \etal~\cite{Drost.2010}, is generated. This set consists of the Euclidean distance $d$ between the two points, the relative angle $\psi$ between their velocity vectors and the individual angles between the velocity vector of the points and their connecting line, $\gamma_{\nu}$ and $\gamma_{u}$. An overview of the various node and edge features for the different data representations is presented in \cref{tab:features}.

\begin{table}[]
	\centering
	\begin{tabular}{@{}llr@{}}
		\toprule
		Invariance          & Node features                       & Edge features                        \\
		\midrule
		-                   & $x, y, v_x, v_y, rcs, t, \text{c}$  & -                                    \\
		Trans.              & $v_x, v_y, rcs, t, \text{c}$        & $dx, dy$                             \\
		Trans. and rot.     & $v, rcs, t, \text{c}$               & $d, \psi, \gamma_{\nu}, \gamma_{u}$  \\
		\bottomrule
	\end{tabular}
	\caption{Set of node and edge features of the graph for different transformation invariant data representations.}
	\label{tab:features}
\end{table}

The matrix transformation maps the previously generated graph to a more easily processable matrix data representation. Therefore, the graph $\mathcal{G}$ is mapped to a tuple consisting of an adjacency matrix $\mathbf{A}$, a node feature matrix $\mathbf{X}$ and an edge feature matrix $\mathbf{E}$. The resulting tuple $(\mathbf{A}, \mathbf{X}, \mathbf{E})$ represents the actual input of the subsequently defined neural network.

\begin{table*}[t]
	\centering
	\begin{tabular}{@{}m{40mm}lrrrrrrr@{}}
		\toprule
		Method                            & Split           & $\text{AP}_{\text{ped}}$ & $\text{AP}_{\text{grp}}$ & $\text{AP}_{\text{tw}}$ & $\text{AP}_{\text{car}}$ & $\text{AP}_{\text{trk}}$ & $\text{mAP}\!\uparrow$   \\
		\midrule
		PointPillars~\cite{Scheiner.2021} & val.           & 11.2           & 22.5            & 38.8            & 54.2             & 53.7             & 36.1           \\
		YOLOv3~\cite{Scheiner.2021}       & val.           & \textbf{34.4}  & 55.7            & 57.4            & 70.2             & 61.9             & 55.9           \\
		RadarGNN (ours)                   & val.           & 34.0           & \textbf{58.3}   & \textbf{66.6}   & \textbf{72.0}    & \textbf{70.1}    & \textbf{60.2}  \\
		\bottomrule
		RadarGNN (ours)                   & test           & 33.1           & 54.8            & 62.0            & 70.4             & 62.0             & 56.5           \\
		\bottomrule
	\end{tabular}
	\caption{Object detection results on the RadarScenes dataset for the translation invariant model configuration. The first three rows are evaluated on the validation (val.) set, while the bottom row is evaluated on the test set. The detection quality is given by the average precision (AP) values for the pedestrian (ped), pedestrian group (grp), two-wheeler (tw), car and large vehicle (trk) class with an IOU threshold of \SI{0.3}{}. The mAP represents the mean average precision over all five classes. All benchmark results can be found in~\cite[Tab.~3]{Scheiner.2021}.}
	\label{tab:radarscenes_obj}
\end{table*}

\subsection{Graph Neural Network Architecture}
\label{sec:gnn}

The graph neural network consists of two major components and is responsible for the generation of an expressive feature representation for the connected detection heads. The two components are the initial feature embedding and the GNN layers themselves.

The initial feature embedding creates a high-dimensional non-contextual feature representation from the low-dimensional node and edge features. Therefore, a shared multilayer perceptron (MLP) with four layers is used for node feature embedding and one with three layers for edge feature embedding.

A major contribution of this paper is the generalization of the previously proposed graph neural network layer of Svenningsson \etal~\cite{Svenningsson.2021}, which is based on Shi \etal~\cite{Shi.2020}. Their graph convolutional layer function

\begin{equation}
	\text{h}^{l+1}_{\nu} = \zeta(\text{h}^{l}_{\nu}, \oplus_{u \in \mathcal{N}_{\nu}}~\xi(x_{\nu} - x_{u}, \text{h}^{l}_{u})) + \text{h}^{l}_{\nu}\text{,}
	\label{eq:gc}
\end{equation}

\noindent updates the node features~$\text{h}_{\nu}$ of node~$\nu$ by adding the results of the update function~$\zeta$ to the input node features~$\text{h}^{l}_{\nu}$ of the current layer~$l$. The update values are determined by an aggregation~$\oplus$ over the node's neighborhood~$\mathcal{N}_{\nu}$ and implemented as a max pooling over the neighbor node features~$\text{h}^{l}_{u}$ weighted by their relative position~$x_{\nu} - x_{u}$. However, this layer formulation does not allow feature dimension changes across GNN layers and only achieves a translational invariance.

To overcome these limitations we propose a more general message passing neural network (MPNN) layer formulation instead of the graph convolutional (GC) layer formulation of Svenningsson \etal~\cite{Svenningsson.2021}. To achieve this, two major changes are made to the update function in \cref{eq:gc}. Firstly, the current node features~$\text{h}^{l}_{u}$ are introduced in the embedding function $\xi$ to allow dimensional changes across GNN layers. Secondly, the possibility of using arbitrary edge features is introduced instead of only using the relative position between neighboring nodes. Therefore, the achievement of certain invariances is subject to the edge and node feature generation rather than the layer formulation itself, which allows the analysis of different invariances without changing the network architecture. The resulting update function~$\zeta$ is given by

\begin{equation}
	\text{h}^{l+1}_{\nu} = \zeta(\text{h}^{l}_{\nu}, \oplus_{u \in \mathcal{N}_{\nu}}~\xi(\text{h}^{l}_{\nu}, \text{h}^{l}_{u}, \text{e}^{l}_{\nu, u}))
	\label{eq:mpnn}
\end{equation}

\noindent where the new node features~$\text{h}^{l+1}_{\nu}$ of node~$\nu$ after layer~$l$ are a function of the original node features~$\text{h}^{l}_{\nu}$ and an aggregation~$\oplus$ over the neighborhood~$\mathcal{N}_{\nu}$. In this process, the aggregation is conducted over the embedding $\xi$ of the sender node features $\text{h}^{l}_{\nu}$, the receiver node features $\text{h}^{l}_{u}$ and the features of the edge $\text{e}^{l}_{\nu, u}$ connecting them. To obtain a permutation invariant network architecture, the update function $\zeta$ and the embedding function $\xi$ are implemented as shared MLPs, and the aggregation function determines the maximum of the embedded features - similar to~\cite{Qi.2017}.

The resulting feature space of these layers forms the foundation for the subsequent detection heads to accomplish the desired perception tasks.

\subsection{Object Detection and Semantic Segmentation}

The proposed RadarGNN model not only performs object detection but also semantic segmentation on the given radar point cloud. This multi-task learning approach is realized by a distinct feature extraction module and individual detection heads, which could also be extended to support additional perception tasks. For our purpose, a semantic segmentation and object detection (bounding box prediction) head is used.

The semantic segmentation head consists of a shared MLP, with a final softmax activation function and predicts a confidence score for each class. The final class for every individual point is then determined by the highest confidence score among all classes.

The object detection head is realized by a shared MLP with two consecutive layers and a linear activation function in order not to restrict the output space. The module predicts a bounding box for every point within the given point cloud, which is why a number of suppression schemes have to be applied to receive the final output. Firstly, a background removal is applied to remove all bounding boxes associated with the background class. Secondly, a non-maximum suppression (NMS) is used to remove all overlapping bounding boxes and keep only the one with the highest confidence score. Thirdly, a class-specific threshold is applied to discard all remaining bounding boxes below a certain confidence. Finally, the absolute bounding box representation is restored.

The overall model is trained with a combined loss function, consisting of multiple task-specific loss functions. The semantic segmentation branch uses a class-weighted cross-entropy loss function $\mathcal{L}_{\text{seg}}$, whereas the object detection branch utilizes a Huber loss function~$\mathcal{L}_{\text{obj}}$~\cite{Huber.1992} with a delta value of one. Additionally, an L2 regularization term $\mathcal{L}_{\text{reg}}$ is introduced to prevent the model from overfitting~\cite{Svenningsson.2021}. The overall loss function is given by

\begin{equation}
	\mathcal{L} = \alpha\mathcal{L}_{\text{seg}} + \beta\mathcal{L}_{\text{obj}} + \gamma\mathcal{L}_{\text{reg}}
	\label{eq:loss}
\end{equation}

\noindent where the weights~$\alpha=1,~\beta=0.5~\text{and}~\gamma=\SI{5e-6}{}$ balance between the different loss terms.

\section{Experimental Results}
\label{sec:results}

The RadarGNN model is evaluated on the RadarScenes dataset and the obtained results are compared to the current state of the art in radar-based object detection and semantic segmentation. In addition, a detailed analysis of the effects of different invariances on the model's performance is given. All experiments are conducted on a dedicated benchmark server and within a containerized environment to keep the evaluation environment constant.

\subsection{Object Detection}

The object detection quality of the RadarGNN model is measured by the mean average precision (mAP) value as defined in~\cite{Scheiner.2021} and with an intersection over union (IOU) threshold of \SI{0.3}{}. In addition, the class-specific average precision (AP) values are reported for the pedestrian (ped), pedestrian group (grp), two-wheeler (tw), car and large vehicle (trk) classes. The RadarGNN method achieves a mAP value of \SI{60.2}{\percent} on the RadarScenes validation set, as shown in \cref{tab:radarscenes_obj}. The highest AP values are achieved for the two-wheeler, car and large vehicle classes, while the lowest AP value is reported on the pedestrian class. To put that into perspective, the results are compared to the current state of the art on the RadarScenes dataset.

Since all previous results for rotated bounding boxes are only reported on the validation set and the source code of none of the comparison models is publicly available, the RadarGNN model is compared to the literature values accomplished on the validation set. However, the results of our model are also reported on the independent test set in the bottom row of \cref{tab:radarscenes_obj}.

The RadarGNN model achieves state-of-the-art results on the RadarScenes dataset and outperforms all previous object detection methods, as shown in \cref{tab:radarscenes_obj}. Our graph-based architecture achieves the highest mean average precision (mAP) value and outperforms the hybrid PointPillars as well as the grid-based YOLOv3 method of Scheiner \etal~\cite{Scheiner.2021}.

The increased object detection quality has multiple causes, although the utilization of point relationships and the preservation of the structural information are of particular importance. Investigations demonstrated that a higher connectivity (number of edges) results in a better object detection score, but consequently in an increase in computational resources. In addition, the results of Scheiner \etal~\cite{Scheiner.2021} indicate that the detection quality of the grid-based approach is limited in respect of the yaw angle prediction because of the discretization and associated loss of structural information, which are not observed with the graph-based approach. Nevertheless, the previous YOLOv3~\cite{Scheiner.2021} method achieves better results for the pedestrian (ped) class, which is characterized by having very few radar points. Additionally, the introduced invariances greatly influence the detection quality, which is discussed below in \cref{subsec:invariance_results}.

To provide more context to these results, the RadarGNN model is also benchmarked in the official nuScenes detection challenge and achieves a nuScenes detection score (NDS) of \SI{0.059}{}. The proposed method, therefore, outperforms the comparison model of Svenningsson \etal~\cite{Svenningsson.2021}, which achieved an NDS of \SI{0.034}{} and has currently the second highest score among all radar-only object detection methods. However, it must be noted that our model was not designed for 3D object detection on the nuScenes dataset but rather for the bird's-eye view (BEV) object detection on the RadarScenes dataset.

\subsection{Semantic Segmentation}

In addition to the object detection quality, the semantic segmentation quality of the proposed multi-task neural network is evaluated on the RadarScenes dataset and compared to the current state of the art. The segmentation quality is measured by the macro-averaged F1 score, where the RadarGNN model achieves a score of \SI{77.1}{\percent} on the RadarScenes test dataset, as shown in \cref{tab:radarscenes_seg}.

\begin{table}[t]
	\centering
	\begin{tabular}{@{}m{50mm}r m{20mm}@{}}
		\toprule
		Method                                          & $\text{F}_{\text{1}}\!\uparrow$   \\
		\midrule
		PointPillars~\cite{Scheiner.2021}               & 47.6                              \\
		YOLOv3~\cite{Scheiner.2021}                     & 53.0                              \\
		LSTM~\cite{Schumann.2017}                       & 59.7                              \\
		PointNet++~\cite{Schumann.2018.2}               & 74.3                              \\
		Recurrent PointNet++~\cite{Schumann.2020}       & 75.0                              \\
		RadarGNN (ours)                                 & \textbf{77.1}                     \\
		\bottomrule
	\end{tabular}
	\caption{Semantic segmentation results on the RadarScenes test set, given by the macro-averaged F1 score.}
	\label{tab:radarscenes_seg}
\end{table}

\begin{figure}[t]\vspace{-4.0mm}
	\centering
	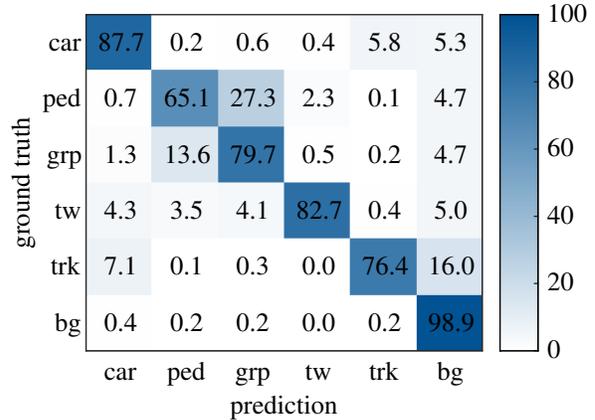\vspace{-2.0mm}
	\caption{Confusion matrix of the semantic segmentation results on the RadarScenes test set. The matrix represents the ground truth values in contrast to the model prediction for the five objects classes and the background (bg) class.}
	\label{fig:confusion}
\end{figure}

The proposed method outperforms all previous radar-based multi-task learning approaches and even the dedicated semantic segmentation models of Schumann \etal~\cite{Schumann.2020}. As shown in \cref{fig:confusion}, the model is able to differentiate between the different classes, while the highest confusion exists between the pedestrian (ped) and pedestrian group (grp) classes. This result indicates the potential of the graph-based approach for additional computer vision applications on radar point clouds.

On the nuScenes dataset a macro-averaged F1 score of \SI{19.6}{\percent} can be achieved on the validation set, which represents the first reported semantic segmentation result with the official nuScenes configuration. Although Nobis \etal~\cite{Nobis.2021} developed a semantic segmentation method on nuScenes radar data, they used a simplified class configuration and achieved a macro-averaged F1 score of \SI{22.8}{\percent}. Using the exact same class configuration as Nobis \etal~\cite{Nobis.2021}, our model achieves a macro-averaged F1 score of \SI{39.4}{\percent}.

\subsection{Transformation Invariances}
\label{subsec:invariance_results}

The proposed approach to transformation invariance allows a detailed analysis of the effects of different invariances on the perception performance. The conducted study compares the object detection and semantic segmentation quality of three models with a non-invariant, translation invariant as well as translation and rotation invariant configuration.

For the object detection task, the highest mAP value of \SI{56.5}{\percent} can be achieved by the translation invariant configuration, as shown in \cref{tab:invariances}. Hence, the non-invariant as well as the translation and rotation invariant configuration accomplished a lower detection score of \SI{19.4}{\percent} and \SI{19.6}{\percent}, respectively.

This result could be caused by both the restriction of the input features as well as the differences in the bounding box description (as described in \cref{sec:radargnn}). For this reason, a complementary study, with a translation and rotation invariant bounding box definition for all three configurations, was conducted. The result of this study shows that the more complex bounding box definition leads to an overall lower detection quality but the general trend remains the same. Within this study the non-invariant and translation invariant methods achieved a mAP of \SI{16.4}{\percent} and \SI{20.8}{\percent}, respectively. Consequently, the input feature restriction can be identified as the root cause of the reduced detection quality. In summary, translation invariance increases the detection quality but the additional rotation invariance, and the accompanying restriction of the input features, negatively affects the detection quality.

Similar to the object detection task, the highest semantic segmentation score of \SI{77.1}{\percent} can be achieved by the translation invariant configuration, as shown in \cref{tab:invariances}. The non-invariant configuration achieved a macro-averaged F1 score of \SI{68.2}{\percent} and the translation and rotation invariant configuration achieved \SI{76.5}{\percent}, respectively.

\begin{table}[t]
	\centering
	\begin{tabular}{@{}m{40mm}rr m{15mm} m{15mm}@{}}
		\toprule
		Invariance                   & $\text{mAP}$   & $\text{F}_{\text{1}}$            \\
		\midrule
		-                            & 19.4           & 68.1                             \\
		Translation                  & \textbf{56.5}  & \textbf{77.1}                    \\
		Translation and rotation     & 19.6           & 76.5                             \\
		\bottomrule
	\end{tabular}
	\caption{Object detection and semantic segmentation results for different invariances on the RadarScenes test set, given by the mean average precision (mAP) and macro-averaged F1 score.}
	\label{tab:invariances}
\end{table}

The smaller differences between the semantic segmentation results, in comparison to the object detection results, can be explained by two reasons. Firstly, the bounding box description has no influence on the results, since semantic segmentation requires no bounding boxes. Secondly, directional information (which is lost with the addition of rotational invariance) is less relevant for point-wise classification than it is for the prediction of the bounding box orientation. 

However, the object detection and semantic segmentation tasks are not independent but coupled by a combined loss function \cref{eq:loss} and model training. Therefore, the segmentation quality is affected by the object detection performance and vice versa. Since the differences between the segmentation results are small, we conducted a complementary study to further analyze the effects of invariances on the segmentation quality. Within this study, the segmentation branch was trained independently by setting $\beta = 0$. As a result, the non-invariant model achieved an F1 score of \SI{60.5}{\percent}, the translation invariant model achieved \SI{66.5}{\percent} and the translation and rotation invariant configuration achieved \SI{68.2}{\percent}. Consequently, the assumption is made that further invariances increase the segmentation quality and justify the necessary restriction of the input features.

\begin{figure}[t]
	\centering
	\def\svgwidth{\linewidth}
	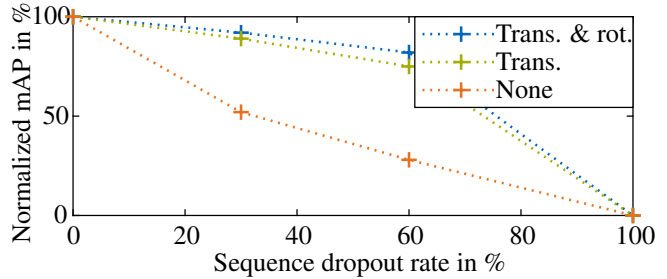
	\caption{Performance of the RadarGNN model for different invariance levels over the number of training sequences. The performance is normalized to the baseline performance.}
	\label{fig:datadrop}
\end{figure}

To provide evidence to the claim that the introduction of transformation invariances counteract the effects of limited data availability, we studied the model performance during a reduction of the training data. For this experiment we gradually reduced the amount of training data sequences, but kept the test set constant and monitored the model performance for the three different levels of invariance. The result shows that transformation invariant models are less affected by a limited data availability and the addition of further invariances contribute positively to this effect as shown in \cref{fig:datadrop}.

As a result, the analysis indicates that certain perception tasks benefit differently from specific transformation invariances. Where semantic segmentation improves with the addition of further invariances, the highest object detection score can be achieved by the translation invariant configuration. Furthermore, additional transformation invariances improve the ability of the model to handle limited data availability.

\section{Conclusion}
\label{sec:summary}

In this paper, we present a graph neural network for both multi-class object detection and semantic segmentation on radar point cloud data. The proposed RadarGNN model uses a generalized message passing neural network layer to consider edge features within its update function and to allow dimensional changes in the GNN. Furthermore, a more generalized approach to achieve transformation invariance is proposed by the creation of an invariant data representation rather than an invariant model architecture. This modification allows the analysis of different invariances without changing the model architecture itself and is transferable to different applications. However, since an invariant data representation always involves a restriction of the input features, a distinct set of point-pair features is proposed to compensate for this during the edge feature generation. The proposed RadarGNN model achieves state-of-the-art results on the RadarScenes dataset for both radar-based object detection and semantic segmentation. In addition, the effects of different invariances on the object detection and semantic segmentation quality is investigated. The incorporation of a sensor fusion concept or the transfer to different sensor modalities is subject to future research.

{\small
	\bibliographystyle{ieee_fullname}
	\bibliography{references}
}

\end{document}